 \documentclass[pmlr,twocolumn,10pt]{jmlr} % W&CP article

\usepackage{booktabs}
\usepackage{multirow}

\usepackage{siunitx}
\usepackage{float}

\newcommand{\equal}[1]{{\hypersetup{linkcolor=black}\thanks{#1}}}

\jmlrvolume{225}
\jmlryear{2023}
\jmlrsubmitted{LEAVE UNSET}
\jmlrpublished{LEAVE UNSET}
\jmlrworkshop{Machine Learning for Health (ML4H) 2023} % W&CP title

 \title[Reinforcement learning agents in N-of-1 trials]{Designing and evaluating an online reinforcement learning agent for physical exercise recommendations in N-of-1 trials}

\author{%
\Name{Dominik Meier}\nametag{$^1$} %\equal{Corresponding author} 
\Email{dominik.meier@student.hpi.de}\\
\Name{Ipek Ensari}\nametag{$^2$}%\footnotemark[1] 
\Email{ipek.ensari@mssm.edu}\\
\Name{Stefan Konigorski}\nametag{$^{1,2,3}$}\equal{Corresponding author} \Email{stefan.konigorski@hpi.de}\\
\AND
\addr $^1$ Hasso Plattner Institute for Digital Engineering, University of Potsdam, Potsdam, Germany\\
\addr $^2$ Hasso Plattner Institute for Digital Health at Mount Sinai, Icahn School of Medicine at Mount Sinai, New York, NY, USA\\
\addr $^3$ Department of Statistics, Harvard University, Cambridge MA, USA\\
}

\begin{document}

\maketitle

\begin{abstract}
Personalized adaptive interventions offer the opportunity to increase patient benefits, however, there are challenges in their planning and implementation. Once implemented, it is an important question whether personalized adaptive interventions are indeed clinically more effective compared to a fixed gold standard intervention.
In this paper, we present an innovative N-of-1 trial study design testing whether implementing a personalized intervention by an online reinforcement learning agent is feasible and effective. Throughout, we use a new study on physical exercise recommendations to reduce pain in endometriosis for illustration. We describe the design of a contextual bandit recommendation agent and evaluate the agent in simulation studies. The results show that, first, implementing a personalized intervention by an online reinforcement learning agent is feasible. Second, such adaptive interventions have the potential to improve patients' benefits even if only few observations are available. As one challenge, they add complexity to the design and implementation process. 
In order to quantify the expected benefit, data from previous interventional studies is required. We expect our approach to be transferable to other interventions and clinical interventions.

\end{abstract}

\section{Introduction}
Chronic diseases are frequently associated with between-patient variability in their symptomatic manifestations and other clinical features. Therefore, one-size-fits-all interventions and strategies are not ideal and yield heterogeneous results. This has led to a shift in study designs in recent years focusing on patients' individual responses, like N-of-1 trials~\citep{Lillie_Patay_Diamant_Issell_Topol_Schork_2011}. N-of-1 trials are multi-crossover randomized controlled trials within one individual that in their classic set-up compare two interventions applied in a pre-determined crossover sequence. Then, individual intervention effects can be estimated and if a series of N-of-1 trials is performed with multiple participants, population-level intervention effects can also be estimated~\citep{Zucker_Schmid_McIntosh_D’Agostino_Selker_Lau_1997}. This approach can be particularly promising for chronic conditions without an established cure or treatment, making patient self-management an important disease management component.
%Additionally, the rising trend of precision medicine aims to individualize treatments for patients through the use of, e.g., pharmacodiagnostic tests~\cite{Jørgensen_2008}.
While N-of-1 trials are the gold standard study design for estimating individual causal intervention effects, comparing more than two interventions can be inefficient as each additional intervention condition linearly extends the total study duration. In recent work, models for adaptive N-of-1 trials have been proposed where the intervention allocation over time can depend on intermediate analyses of the trial~\citep{Senarathne_Overstall_McGree_2020, Malenica2021, Shrestha_Jain_2021}. 

\textbf{Contributions.} 
As a first contribution, we describe an innovative novel study design that evaluates in an N-of-1 trial whether a personalized intervention is superior compared to a generic intervention, where the personalized intervention is itself an adaptive reinforcement learning (RL) agent which learns the best of multiple interventions in a given context. To the best of our knowledge, incorporating an RL agent as one intervention in an N-of-1 trial has not been reported before. 
Second, we describe the process of designing and evaluating an RL agent tailored to this study design.
For this, we use a trial for illustration that investigates physical exercise recommendations to reduce pain in patients with endometriosis.
A priori, it is not clear if a personalized RL agent is able to learn in a setting such as our presented pilot study.
Our results show that implementing a personalized intervention by an online reinforcement learning agent is both feasible and can be effective, even if only few data points are available and learning is done on the individual level without pooling across participants.

Endometriosis is an estrogen-mediated inflammatory disease characterized by the growth of endometrial-like tissue outside the uterus, leading to painful adhesions and lesions. It is associated with debilitating pelvic pain and infertility, and a 7-year delay in diagnosis, in part due to significant between-patient symptomatic variability~\citep{Tamaresis_Irwin_Goldfien_Rabban_Burney_Nezhat_DePaolo_Giudice_2014, Schliep_Mumford_Peterson_Chen_Johnstone_Sharp_Stanford_Hammoud_Sun_BuckLouis_2015, Fourquet_Zavala_Missmer_Bracero_Romaguera_Flores_2019}. Despite recent efforts, endometriosis remains poorly understood and is not well-managed, and there is a critical need to identify novel methods for disease management with a focus on pain reduction~\citep{Rogers_DHooghe_Fazleabas_Gargett_Giudice_Montgomery_Rombauts_Salamonsen_Zondervan_2009}. There is promising evidence on physical exercise for endometriosis pain management, however, it is unclear which physical activity with which intensity and duration is best~\citep{Bonocher_Montenegro_RosaeSilva_Ferriani_Meola_2014}. In this setting, using adaptive methodology, which allows to directly incorporate individual responses into further intervention decisions, offers a great potential to increase the efficiency of the trial. 
Here, we combine the N-of-1 trial design with an adaptive intervention arm, designed to maximize patient's benefit. In the adaptive arm, personalized physical exercise recommendations are given by an online RL agent.

The rest of the paper is structured as follows: 
In \autoref{sec:related-work}, we reference previous studies on exercise and pain in endometriosis and give background for our RL approach.
In \autoref{sec:causal-effects}, we present a conceptual causal model of our assumed effects of exercise on endometriosis-related pain that serves as rationale for our selected interventions and study design and informs the design of the RL agent. \autoref{sec:study-design} describes our study setup in detail.
\autoref{sec:requirements} describes the requirements for the usage of an agent to select intervention decisions in this scenario, and \autoref{sec:agent-design} explains our chosen architecture, followed by an evaluation of the agent in simulation studies described in \autoref{sec:evaluation}. We conclude with a discussion in \autoref{sec:discussion}.

\section{Related work}
\label{sec:related-work}
\subsection{Exercise interventions in endometriosis}

Despite its prevalence of 1 in 10 women being affected worldwide, endometriosis is poorly managed and without well-established treatment or cure~\citep{Cramer_Missmer_2002, Jarrell_Mohindra_Ross_Taenzer_Brant_2005, Stratton_Sinaii_Segars_Koziol_Wesley_Zimmer_Winkel_Nieman_2008, nhcp_estimates}. Current medical intervention efforts in endometriosis focus on pharmacological symptom management, which is associated with unpleasant side effects and inadequate efficacy%and does not provide adequate relief in the long term
~\citep{Rogers_DHooghe_Fazleabas_Gargett_Giudice_Montgomery_Rombauts_Salamonsen_Zondervan_2009,
Sinatra_2010, Krebs_Gravely_Nugent_Jensen_DeRonne_Goldsmith_Kroenke_Bair_Noorbaloochi_2018}. %Hysterectomies, for more persistent cases, are associated with a return of the painful symptoms in one third of the patients. 
Patients report a variety of strategies for symptom self-management. % ~\citep{Ensari_Pichon_Lipsky-Gorman_Bakken_Elhadad_2020, Armour_Sinclair_Chalmers_Smith_2019}. 
Of these, physical exercise remains the most promising and is associated with consistently favorable effects~\citep{
Gonçalves_Barros_Bahamondes_2017,
Armour_Ee_Naidoo_Ayati_Chalmers_Steel_Manincor_Delshad_2019,
Armour_Sinclair_Chalmers_Smith_2019,
Ensari_Lipsky-Gorman_Horan_Bakken_Elhadad_2022}. 
However, data are scarce and suggest variability across patients with respect to exercise preferences and contextual factors that influence efficacy~\citep{
Bonocher_Montenegro_RosaeSilva_Ferriani_Meola_2014,
Ricci_Viganò_Cipriani_Chiaffarino_Bianchi_Rebonato_Parazzini_2016,
Belavy_VanOosterwijck_Clarkson_Dhondt_Mundell_Miller_Owen_2021}. %, Ensari_Lipsky-Gorman_Horan_Bakken_Elhadad_2022}.
Previous work further indicates that regular exercisers are more likely to experience decreased pain responses after %(i.e., reduction or prevention) subsequent to acute bouts of 
exercise bouts and that a wide range of modalities are utilized, including aerobic, calisthenic, and stretching-oriented activities ~\citep{Ensari_Lipsky-Gorman_Horan_Bakken_Elhadad_2022}.

% Classification of physical exercise in type, intensity, duration.
For building interventions in our study, we follow the standard exercise prescription method within clinical exercise physiology which considers 4 components: type, duration, intensity, frequency~\citep{Franklin_2021}. Physiologists typically update the individual's prescription at each follow-up appointment based on the individual's responses. We propose to automate this learning and updating process using RL.

\subsection{Contextual bandits in adaptive trials}
Adaptive designs allow responding to collected data quickly and efficiently
~\citep{Pallmann_Bedding_Choodari-Oskooei_Dimairo_Flight_Hampson_Holmes_Mander_Odondi_Sydes_2018}.
Having learned adaptive sequences, Just-in-Time Adaptive Interventions can be developed which provide interventions to patients based on contextual features~\citep{Nahum-Shani_Smith_Spring_Collins_Witkiewitz_Tewari_Murphy_2018}.
In the adaptive recommendations' field, studies have mainly focused on investigating which personalized messaging can help to improve the proximal outcome.
For example, the HeartSteps project used an RL algorithm based on Thompson Sampling to personalize notifications for sedentary patients~\citep{Liao_Greenewald_Klasnja_Murphy_2020}.
In another study, Boltzmann sampling was used to personalize messages for sedentary patients with diabetes~\citep{Yom-Tov_Feraru_Kozdoba_Mannor_Tennenholtz_Hochberg_2017}.
In the SleepBandits project, patients had the possibility to self-experiment with different interventions on their sleep. Recommendations were provided based on Thompson Sampling, which led to increased sampling of beneficial conditions~\citep{Daskalova_Yoon_Wang_Araujo_Beltran_Nugent_McGeary_Williams_Huang_2020}.

\section{Conceptual model for pain modulation via exercise}
\label{sec:causal-effects}
\begin{figure}[h]
    \centering
    \includegraphics[width=0.8\linewidth]{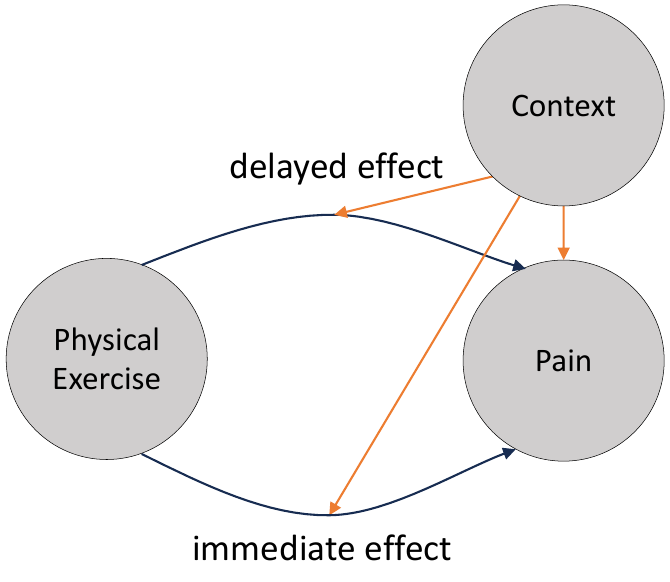}
    \caption{Causal graph with immediate and delayed effects. Context moderates effect on pain but also influences pain.}
    \label{fig:causal-diagramm}
\end{figure}

The overall goal of this study is to design an RL agent that can efficiently identify the optimal dose (w.r.t. type, intensity, duration) of exercise for pain reduction for a given individual. To achieve this, the RL agent needs to learn quickly with limited data and determine the optimal exercise intervention per individual. This requires consideration of the temporality of the pain modulating effects of exercise based on expert knowledge. Exercise has both acute (i.e., immediate) and delayed (i.e., distal) effects acting through various possible pathways, and we depict these timelines and our assumed causal graph in \autoref{fig:causal-diagramm}. We focus on immediate effects. Nevertheless, we give a brief overview of all effects in the following. 

\subsection{Immediate effects of exercise on pain}
There are various psychological and physiological factors that can explain the immediate (i.e., lasting up to several hours post-exercise bout) pain modulating effects of exercise in endometriosis. 
\paragraph{Psychological effects}
Psychological factors such as a sense of accomplishment and/or self-efficacy, accompanied by an overall positive mood state, can favorably modulate post-exercise pain responses~\citep{Middelkamp_van_Rooijen_Wolfhagen_Steenbergen_2017}. Similarly, the distraction hypothesis of exercise has been demonstrated for pain and related outcomes~\citep{Nolen-Hoeksema_Morrow_Fredrickson_1993,
Villemure_Bushnell_2002,
Mikkelsen_Stojanovska_Polenakovic_Bosevski_Apostolopoulos_2017}.
\paragraph{Physiological effects}
Evidence suggests acute improvements in different types (e.g., pressure, thermal, cold, intensity) of pain sensitivity post-exercise~\citep{Naugle_Fillingim_Riley_2012}, accompanied by changes in immediate hormone profile \citep{West_Phillips_2012} and autonomic responses (e.g., reduced pain receptor reactivity)~\citep{Uzawa_Akiyama_Furuyama_Takeuchi_Nishida_2023}. Similarly, some types of pain (e.g., nociceptive, somatic) can respond favorably to muscular relaxation and enhanced movement in the pain-impacted  area~\citep{Louw_Zimney_Puentedura_Diener_2016}. As endometriosis pain often originates from multiple body areas, the type of the pain varies~\citep{Ensari_Pichon_Lipsky-Gorman_Bakken_Elhadad_2020}.  
\subsection{Delayed effects of exercise on pain}
Besides immediate effects, there is evidence for delayed effects of exercise on chronic pain.
\paragraph{Improved physical function}
A wider range of musculoskeletal and other physiological improvements are expected in the long term with exercise, which subsequently can improve pain sensitivity and reactivity and associated quality of life~\citep{Ambrose_Golightly_2015, Rice_Nijs_Kosek_Wideman_Hasenbring_Koltyn_Graven-Nielsen_Polli_2019} 
\paragraph{Improved endocrine function}
Especially relevant for endometriosis, there is strong evidence for exercise-mediated endocrine system, particularly estrogen and other sex hormones, for pain modulation in menstrual pain disorders~\citep{Jaleel_Shaphe_Khan_Malhotra_Khan_Parveen_Qasheesh_Beg_Chahal_Ahmad_etal._2022}.
\paragraph{Improved inflammatory markers}
Different exercise intensities and modalities have been shown to improve inflammatory marker profiles~\citep{Bote_Garcia_Hinchado_Ortega_2013} and inflammatory responses~\citep{Athanasiou_Bogdanis_Mastorakos_2023}. This is of particular relevance to endometriosis given its inflammatory nature.  
\paragraph{Other neuromodulator adaption}
Existing evidence suggests that physical activity modulates central nervous system excitability and inhibition, and that chronic exercise can downregulate endogenous pain pathways~\citep{Sluka_Frey-Law_HoegerBement_2018}.

\subsection{Moderation through context}
We expect moderation of both immediate and delayed effects by the current context of the patient, for which we assess past and current (i) intensity and duration of activities and (ii) endometriosis-related pain.

Because we expect pain measurements over time to be autocorrelated, context also influences pain in our model.
The context variables will be used to personalize the exercise recommendations by the agent. Pain areas can vary across patients and over time. Here, we consider pain in the lower abdominal (including pelvis), groin and hip areas as primary endometriosis-related pain outcomes. As secondary outcomes, we consider other body locations tracked by the patients, which is accomplished through use of a visual body map similar to the visual body map in the McGill Pain Scale~\citep{Melzack_1975} that allows tracking all body areas to indicate pain location, as well as intensity and type of the pain.

\section{Study design of the N-of-1 trial}
\label{sec:study-design}
\begin{figure*}[tb]
    \centering
    \includegraphics[width=\linewidth]{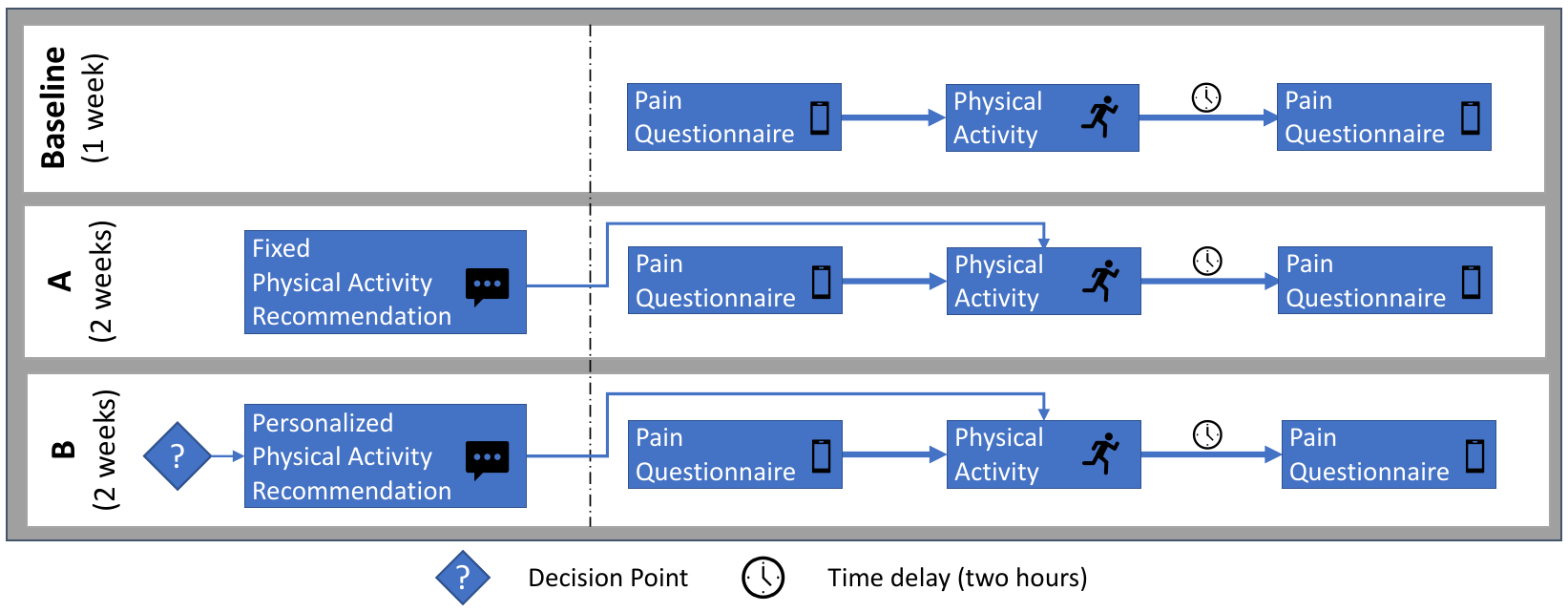}
    \caption{Five weeks randomized pilot study design with Baseline followed by randomized A-B or B-A.\\ In the baseline phase, no recommendations are given, but patients can track exercising. In Phase A, fixed exercise schedules are recommended, which is compared to Phase B, in which personalized exercise recommendations are provided by a Reinforcement Learning agent. The agent uses data points from all previous phases to make a personalized recommendation. Throughout the five weeks, data will be collected using wearables and questionnaires for later analysis.}
    \label{fig:study-design}
\end{figure*}

In this section, we describe the proposed pilot N-of-1 trial, see \autoref{fig:study-design} for a visualization. 
After an initial baseline phase of one week, patients are randomized into an A-B or B-A design, where each A or B phase is 2 weeks long. 
The baseline phase serves to preconfigure the agent using tracked data points. An initial assessment is used to define the set of exercises available for recommendation.
%\paragraph{A Phase}
In Phase A, patients receive generic physical exercise recommendations based on the published PA Guidelines for adults~\citep{pa_guidelines_2018}.
%\paragraph{B Phase}
In Phase B, patients receive RL-generated personalized exercise recommendations.
All exercise recommendations are delivered through an mHealth app, which allows integrating real-time data from patients into the recommendations. 
Of note, while typical physical exercise guidelines assume exercises done 3–5 times per week, in this pilot study we propose daily exercises so that more observations can be made under each intervention, providing an easier setting for the RL agent to learn. 
%In the following, we will explain the different parts of the study in detail:
%\paragraph{Beginning Assessment}
%
%\paragraph{Tracking}
As outcomes, patients will self-track current symptoms (including pain location, type, intensity), daily functioning, health behavior, and exercise in the morning, evening, before, and after their exercise through the app.
The primary outcome for the RL agent is pain measured on a visual analog scale (VAR).
%Symptoms are tracked by asking participant for current pain levels in the morning, evening, before and after their exercise. 
In addition, a weekly PROMIS pain interference questionnaire\footnote{\url{https://cde.nida.nih.gov/sites/nida_cde/files/PROMIS\%20Adult\%20ShortForm\%20v1.0\%20Pain\%20Interference\%206b.pdf}} is administered.
%\paragraph{A Phase}
%In Phase A, patients receive generic physical exercise recommendations based on the published PA Guidelines for adults~\citep{pa_guidelines_2018}.
%\paragraph{B Phase}
%In Phase B, patients receive RL-generated personalized exercise recommendations.
The study has been reviewed and approved by the Mount Sinai IRB Board (STUDY-23-00721).

\section{Requirements for a personalized agent inside an N-of-1 trial}
\label{sec:requirements}

In the following section, we describe requirements for the RL agent. For this, we adapt the PCS framework \citep{Trella_Zhang_Nahum-Shani_Shetty_Doshi-Velez_Murphy_2022} to a personalized RL agent inside an intervention arm for an N-of-1 trial.

\subsection{Personalization (P)}
\label{subsec:personalization}
Few data points need to be sufficient for effective personalization.
Phase B of our pilot study will only include 14 decision points for the agent over two weeks.
These few decision points make efficient personalization challenging.
To overcome this, we use the interventional data from the baseline phase, as well as from the A phase in case of an A-B randomization, for agent pre-configuration.

\subsection{Computability (C)}
\label{subsec:computability}
Model updates and recommendations need to be feasible.
This requires timely collection of contextual information from patients and timely intervention decisions.
This is guaranteed by integrating the agent in an mHealth platform when implementing the study.

\subsection{Stability (S)}
\label{subsec:stability}
The agent may never harm patients. We use the beginning assessment to define a safe subset of possible interventions to ensure this.
Additionally, the agent needs to be robust in real-life scenarios. This includes robustness against model misspecification, missing-data and non-adherence to exercise recommendations.
The pilot format of our study requires that the data generated by the agent are usable for post-trial inference and for informing the implementation of a follow-up study.

\section{Agent design}
\label{sec:agent-design}
In this section, we describe our design decisions for the agent.
To design our agent, we make two simplifying assumptions:
First, based on the description in \autoref{sec:causal-effects}, we assume that the state is fully captured by the context variables and that there is no hidden state. Hence, the context moderates the effect of physical exercise on pain.
Second, we assume that it is sufficient to pick the recommendation which is most effective at the current time point, and disregard long-term planning of recommendations.
Similar assumptions have been made in many mHealth intervention settings (e.g., \cite{Liao_Greenewald_Klasnja_Murphy_2020, Figueroa_Aguilera_Chakraborty_Modiri_Aggarwal_Deliu_Sarkar_JayWilliams_Lyles_2021}), and allow fast learning in few data point environments.
Making these assumptions allows us to model the selection of a recommendation out of the predetermined set as an instance of a contextual multi-armed bandit problem.
We design the RL agent as a contextual bandit using a Thompson sampling policy based on a Bayesian model.
Next, we describe the action space, policy, and reward function of the agent.
\subsection{Action space}
The action space defines which interventions can be selected.
Based on the beginning assessment of patients, we create a personalized set of physical exercise recommendations. Each action is defined by type, duration, and intensity.

\subsection{Policy}
We choose Thompson Sampling as underlying policy of the agent due to its widespread use in adaptive trials (e.g., ~\cite{
Daskalova_Yoon_Wang_Araujo_Beltran_Nugent_McGeary_Williams_Huang_2020,
Liao_Greenewald_Klasnja_Murphy_2020}) and empirically good performance in diverse settings~\citep{Chapelle_Li}.
Thompson Sampling uses the probability of being the optimal intervention in a given context as selection probability of an intervention.
We define the optimal intervention as the intervention with the largest pain reduction, and estimate the probability by fitting a Bayesian model for prediction of the measured pain reduction $\Delta pain$.
We assume that using the difference in pain measurements pre- and post-exercise allows us to only measure the immediate effect of the exercise and reduce autocorrelation in the time series, which we otherwise would have to incorporate in the model.

\subsubsection{Bayesian model}
\label{subsubsec:bayesian-model}
% Bayesian modeling allows the pre-configuration of the agent with priors on the coefficient values, in order to make sure that reasonable recommendations are made from the beginning
% In order to guarantee the sample-efficiency, it will be essential to be able to pre-configure the agent with good priors, that allow giving good recommendations from the beginning. This can include personalization before trial starts, e.g., based on patient characteristics.
We model the pain reduction as a sum of four individual terms, which capture the different assumed effects of exercise type, intensity, duration, and pain.
Let $i=1...n$ denote patients, $j=1...k$ denote the exercise recommendation in the personalized set of physical exercises, and let $t=1...T$ denote the decision points.
Each exercise is characterized by its type, intensity, and duration and we define
\begin{align*}
     \text{Burden}_{ij} = \text{Intensity}_{ij} \cdot \text{Duration}_{ij},
\end{align*}

where exercise intensity is manually assigned a value between 0 and 1 and duration is assessed in minutes and divided by the maximum duration of any exercise. As each exercise type is assigned an intensity, it affects burden through its intensity. %Different exercise types are categorically encoded from $1..k$, and intensity and duration are linearly scaled to the $[0, 1]$ interval.
This allows the agent to understand that some interventions are similar in intensity and duration. %, even if they differ in type.
See \autoref{table:set_of_interventions} for an example.

As inputs for the calculation of the expected pain reduction, we define
\begin{align*}
    pain_{i, t} &= \text{Current pain of patient $i$ at $t$}.\\
    mi_{i, t} &= \text{Mean intensity of last 3 exercises for $i$ at $t$}\\
    md_{i, t} &= \text{Mean duration of last 3 exercises for $i$ at $t$}\\
\end{align*}

Our Bayesian linear regression model includes four summands:
$\tau_{\text{Type},i,j}$ models an intercept for each type of physical exercise per patient.
$y_{\text{Intensity},i,j,t}$ models the effect of the intensity of the physical exercise on pain.
We model this as a linear interaction between the intensity of the proposed physical exercise and the mean intensity of the last three exercises.
Similarly, $y_{\text{Duration},i,j,t}$ models this effect for duration of exercise.
$y_{\text{Burden},i,j,t}$ models an interaction effect between current pain and burden.
We use burden instead of individual coefficients for duration and intensity to decrease the number of parameters.
This gives the following model formula:
\begin{align}
    \Delta pain_{i, j, t} &\sim \text{Normal}(\mu_i, \sigma_i) \label{eq:delta_pain} \\ 
    \mu_i &= \tau_{\text{Type},i,j} + y_{\text{Intensity},i,j,t} \nonumber \\
     &+ y_{\text{Duration},i,j,t} + y_{\text{Burden},i,j,t} \nonumber \\ \nonumber
    y_{\text{Intensity},i,j,t} &= (\alpha_i + \beta_i \cdot mi_{i, t})\cdot \text{Intensity}_{i,j}\\ \nonumber
    y_{\text{Duration},i,j,t} &= (\gamma_i + \delta_i \cdot md_{i, t}) \cdot \text{Duration}_{i,j}\\ \nonumber
    y_{\text{Burden},i,jt} &= (\eta_i + \kappa_i \cdot pain_{i, t})\cdot \text{Burden}_{i,j} \nonumber
\end{align}
\newpage This model has 7 parameters plus the number of different types of activities for each patient. We use $\sigma \sim \text{Exponential}(1)$, and $\alpha, \beta, \gamma, \delta, \eta, \kappa, \tau \sim \text{Normal}(0, 1)$ as non-informative priors.

\subsection{Reward}
We use $\Delta pain_{i, j, t}$ as reward, which is the decrease in pain measured two hours after performing the exercise relative to the pain measured before the exercise on the VAR scale.

\subsection{Implementation}
The model was implemented and fitted using the NUTS Sampler provided by PYMC~\citep{Salvatier_Wiecki_Fonnesbeck_2016}.
To generate the probability of best intervention, we sampled the posterior predictive distribution with the parameters set to the patient's context to obtain an estimated pain reduction.
The relative frequencies of the greatest pain reduction were then used as the selection distribution for Thompson Sampling.

\section{Evaluation}
\label{sec:evaluation}
In this section, we describe the evaluation of the performance of our proposed agent. Our goal is to assess whether we can confidently deploy the agent in the pilot study.
As data from previous studies is not available for evaluating our proposed agent, 
we test robustness and personalization across different scenarios in a simulation study. 

\subsection{Simulation setup}
The evaluation consists of three main steps.
First, we define different scenarios, which each correspond to different assumptions on the environment.
Second, to quantify performance, we translate the PCS requirements (see  \autoref{sec:requirements}) into metrics. 
% All calculated metrics are shown in \autoref{tab:results}.
Third, we simulate the behavior of the agent for 100 patients for each of the seven scenarios and each of the two randomization schemes (A-B, B-A), for a total of 14 simulations, and report the performance of the agent with respect to the defined metrics.
%In the following, we describe each step in detail:

For our simulation, we use a set of eight possible exercise recommendations shown in \autoref{table:set_of_interventions}.
\begin{table}[]
\begin{tabular}{lllll}
Type & Intensity & Duration     & Comment      &  \\
\midrule
0    & 0.3       & 0.5 (30min)  & Slow jogging &  \\
0    & 0.5       & 0.5 (30min)  & Jogging      &  \\
0    & 0.7       & 0.5 (30min)  & Fast jogging &  \\
1    & 1         & 0.1 (6min)   & HIIT         &  \\
1    & 1         & 0.2 (12min)  & HIIT         &  \\
1    & 1         & 0.3 (18min)  & HIIT         &  \\
2    & 0.5       & 0.75 (45min) & Swimming     &  \\
3    & 0.1       & 1 (60min)    & Yoga         & 
\end{tabular}
    \caption{Example set of physical exercise recommendations used for simulation}
    \label{table:set_of_interventions}
\end{table}
The simulation mirrors the study configuration.
The baseline phase and the A phase in case of A-B randomization were simulated using a fixed sequence of these exercise recommendations and used as input data for the agent.
Our simulation did not include missing values.
The Python code and results are available from the GitHub repository\footnote{\url{https://github.com/HIAlab/Reinforcement-learning-agents-in-N-of-1-trials}}.

\subsection{Evaluation scenarios}
In \autoref{sec:causal-effects}, we presented the assumed effects of exercise on pain.
As we have little information about the true effects in the graph in \autoref{fig:causal-diagramm}, %which SEM correctly describes our environment, we test the performance of the agent in 
we generate six different possible scenarios (I-VI) in order to test stability (see \autoref{subsec:stability}) across different environments. An additional Scenario VII is included to test the effects of informative non-adherence: %Our four models describing the environments are defined as follows:
\paragraph{Scenario I}
Pain reduction is drawn independently of other variables from a Normal(0,1) distribution. This scenario is included as a null scenario. %to test the behavior in case learning meaningful effects is not possible.
\paragraph{Scenario II}
Pain reduction is calculated from the linear model specified in \autoref{eq:delta_pain}. All parameters are drawn individually for each patient from a Normal(0,1) distribution, and are then fixed for this patient across the study. This environment is included to test the behavior of the agent when the underlying model is correctly specified. %for a match between SEM and our Bayesian Model.
\paragraph{Scenarios III, IV, V, VI}
Pain reduction in Scenarios III, IV, V, VI is defined as in Scenario II, but sets the parameters regarding type (Scenario III), or intensity (Scenario IV), or duration (Scenario V), or intensity and duration (Scenario VI) to 0. These environments are included to test robustness of the agent against partially violated assumptions of the Bayesian model.

\paragraph{Scenario VII}
To investigate the behavior of the agent when data is missing not at random, we include Scenario VII.
Pain reduction is calculated like in Scenario II, but informative non-adherence using the following rationale is simulated: If the recommendation that is given by the agent would increase the pain (based on the linear model), this data point is dropped in 50\% of the cases, and will not be used as feedback to the agent.

\subsection{Metrics}
We evaluate three key metrics that have similarly been proposed by \cite{Trella_Zhang_Nahum-Shani_Shetty_Doshi-Velez_Murphy_2022} to assess the personalization of an agent:
\begin{itemize}
    \item Average of users' average regret
    \item The 25th percentile of users' average regret
    \item Average regret for multiple time points
\end{itemize}
We define the regret as the improvement over the non-adaptive arm, calculated as the difference between pain reduction under the fixed arm and the adaptive arm simulated with the same context variables:
\begin{align*}
    \text{Regret}_{i,t} = \sum_{l=1}^t \Delta pain_{i, \text{fixed}, l} - \sum_{l=1}^t \Delta pain_{i, \text{adaptive}, l}
\end{align*}

Since our pilot study is meant to inform further research, it is important to generate a dataset useful for post-trial inference.
The greatest threat to meaningful post-trial inference is if some participants only receive the same intervention, due to fast convergence of the Bayesian model.
We measure the diversity of interventions by five different metrics:
For assessing the different types of exercise recommendations (discrete case), we calculate the Shannon entropy~\citep{shannon} as a measure of randomness.
For the continuous variables intensity and duration, we calculate the standard deviation.
Additionally, we calculate the mean of the minimum and maximum probability for intervention selection.

\subsection{Results}
\begin{figure*}[tbp]
    \subfigure[A-B Randomization]{
        \includegraphics[width=0.48\linewidth]{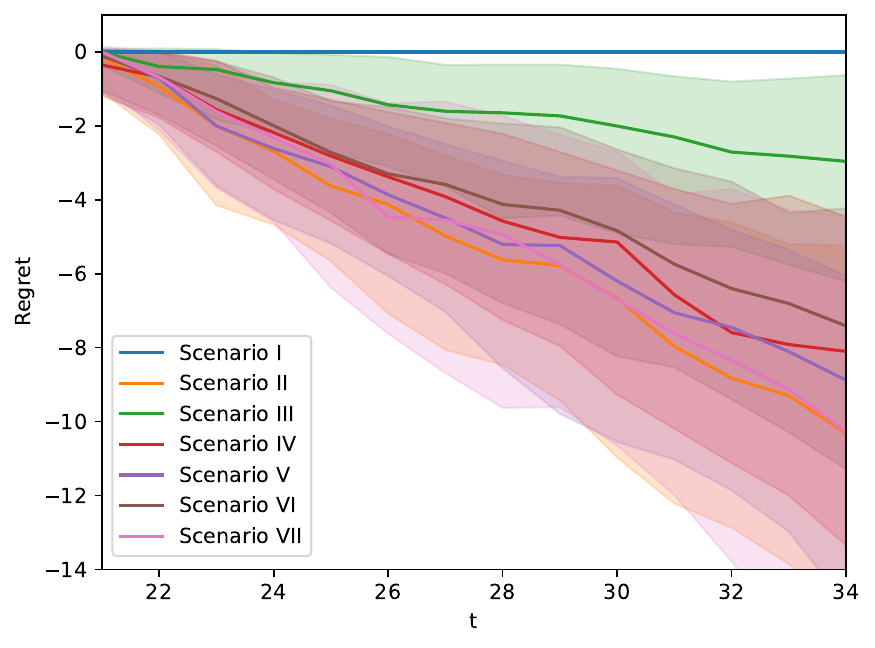}
        \label{subfig:lineplot-regret-ab}
    }
    \subfigure[B-A Randomization]{
        \includegraphics[width=0.48\linewidth]{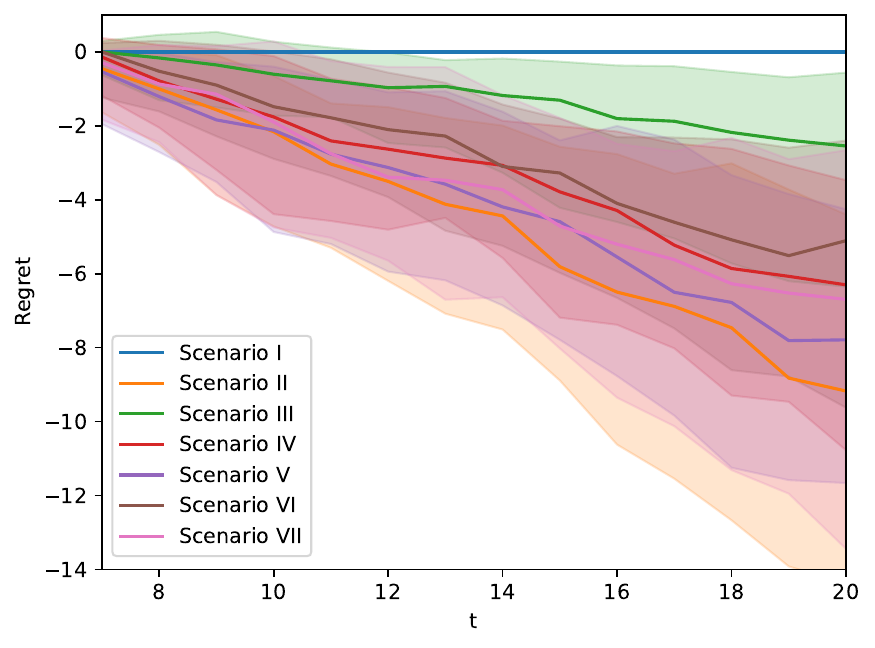}
        \label{subfig:lineplot-regret-ba}
    }
    \caption{Results of simulation study showing the regret over time $t$. Lines show the median regret over time across the 100 simulated patients. Shaded areas show the 25\% and 75\% quantiles.}
    \label{fig:lineplot-regret}
\end{figure*}

As seen in \autoref{tab:results}, and \autoref{fig:lineplot-regret}, the agent reaches a negative regret for all scenarios expect Scenario I. Clear downward trends are visible for the median and quartiles.
The regret in Scenario I is zero, since under the null scenario, the pain reduction is independent of the chosen intervention.

In the B-A randomization, the agent only has seven data points from the baseline to learn the participant's coefficients.
We therefore observe higher regret (i.e., worse performance) compared to A-B randomization.
We observe that regret is slightly negative in the 25\% worst case. This shows that personalization not only works for the average patient but for most patients.
The mean maximum selection probability across all environments is 44\%, while the minimum is at 1\%.
This suggests that low values for regret do not come from selection of good interventions, but rather through discarding bad interventions.

In the non-adaptive arm with the generic intervention, the entropy is 1.23 for exercise type and the standard deviations are 0.26 for duration and 0.32 for type. In the adaptive arm applying the RL agent, we mostly observe smaller values for entropy and standard deviation.
Especially the entropy of exercise type in Scenario II is low with 0.82 in the A-B design. This suggests that a correctly specified model leads to a fast model convergence. At the same time, this will create high beneficence for patients in pain reduction. 
The non-informative adherence in Scenario VII did worsen the 25\% worst case performance slightly, as well as the mean performance in the B-A design. In the A-B design, the agent still performed similar to Scenario II.

\section{Discussion}
\label{sec:discussion}

Here, we have presented an innovative N-of-1 trial study design comparing a personalized intervention by an online RL agent to a generic intervention. We focused on the study design and the design of the RL agent, while leaving the design of post-trial inference to follow-up studies. Regarding the RL agent design, the results of simulation studies show that employing such an RL agent can be feasible also when only few data points are available and learning is done on the individual level without pooling across patients. 

Our proposed agent showed the ability to learn in different scenarios, therefore allowing to create study designs which are flexible across different plausible scenarios.
In this low-sample setting and even when the set of possible interventions is large, we expect benefit from personalization.

Finding the right balance between specialized and general agents remains a challenge.
Specialization can lead to higher optimization potential, e.g., through including prior knowledge, but at the same time might decrease robustness against model misspecification.
Due to the limited data available for our pilot trial, we prioritized robustness and selected a more general agent.

Under the assumption that data is missing completely at random, missing data will slow down the learning of the agent due to less available data points, but would not interfere with the learning of the Bayesian model. Once our pilot study will be completed, we will investigate the behavior of the agent when data is missing not at random in this real data.

Our work has some limitations. For real-world applications, additional challenges such as missing data need to be considered. While we made simplifying assumptions in our simulation study, such challenges might make it necessary to include additional safety measures inside the agent.
Also, due to time constraints for the total length of the study and our main goal of proofing the feasibility of integrating an RL agent inside an N-of-1 trial, we opted for daily exercise recommendations to collect more data points. This is more frequent than usual exercise frequencies in typical studies. However, we expect that our results generalize to longer studies with less frequent exercising, and are planning to reduce the frequency in follow-up studies.
Finally, it will be interesting to investigate model misspecification not only when the model is misspecified (see Scenario I) but also when the priors in the Bayesian model are not well selected.

A natural extension and possible requirement in some situations to improve the learning rate in a low-sample setting is to include pooling, which has been effectively deployed in previous studies. However, it might also worsen personalization in highly heterogeneous environments~\citep{Tomkins_Liao_Klasnja_Murphy_2020, Trella_Zhang_Nahum-Shani_Shetty_Doshi-Velez_Murphy_2022}.
Judging how and to which extent pooling might improve the agent for our scenario is hard to estimate without additional data. Since we already observed benefit for participants without pooling, we decided to disregard it for this pilot study.

We are curious to see the benefit of personalizing agents for pain reduction in endometriosis and beyond, and to observe the increasing use of N-of-1 trials for their evaluation.

\begin{table*}[tbp]
    \centering
   \begin{tabular}{llrrrrrrr}
   \multirow[l]{2}{*}{Scenario} & \multirow[l]{2}{*}{Design} & \multicolumn{2}{c}{Regret} & Entropy & \multicolumn{2}{c}{Standard Deviation} & \multicolumn{2}{c}{Mean Probability} \\
   \cmidrule(lr){3-4} \cmidrule(lr){6-7} \cmidrule(lr){8-9}
   & & Mean & 0.75 quantile & Type & Duration & Intensity & Max & Min\\
\toprule
\multirow[c]{2}{*}{I} & A-B & 0.00 & 0.00 & 1.10 & 0.25 & 0.30 & 0.29 & 0.05 \\
 & B-A & 0.00 & 0.00 & 1.12 & 0.27 & 0.31 & 0.34 & 0.03 \\
\addlinespace[1ex]
\multirow[c]{2}{*}{II} & A-B & -11.76 & -5.24 & 0.82 & 0.22 & 0.25 & 0.44 & 0.01 \\
 & B-A & -10.50 & -4.40 & 0.90 & 0.24 & 0.27 & 0.46 & 0.01 \\
\addlinespace[1ex]
\multirow[c]{2}{*}{III} & A-B & -5.12 & -0.63 & 0.93 & 0.23 & 0.26 & 0.36 & 0.03 \\
 & B-A & -4.40 & -0.56 & 0.98 & 0.26 & 0.28 & 0.39 & 0.02 \\
\addlinespace[1ex]
\multirow[c]{2}{*}{IV} & A-B & -9.98 & -4.46 & 0.85 & 0.22 & 0.25 & 0.43 & 0.01 \\
 & B-A & -8.32 & -3.47 & 0.95 & 0.25 & 0.28 & 0.44 & 0.02 \\
\addlinespace[1ex]
\multirow[c]{2}{*}{V} & A-B & -10.74 & -6.07 & 0.81 & 0.22 & 0.25 & 0.42 & 0.01 \\
 & B-A & -9.48 & -4.26 & 0.90 & 0.24 & 0.27 & 0.44 & 0.01 \\
\addlinespace[1ex]
\multirow[c]{2}{*}{VI} & A-B & -8.76 & -4.23 & 0.86 & 0.23 & 0.27 & 0.42 & 0.01 \\
 & B-A & -7.02 & -2.40 & 0.96 & 0.25 & 0.28 & 0.43 & 0.02 \\
\addlinespace[1ex]
\multirow[c]{2}{*}{VII} & A-B & -11.74 & -4.42 & 0.82 & 0.22 & 0.25 & 0.44 & 0.01 \\
 & B-A & -8.37 & -2.66 & 0.97 & 0.25 & 0.28 & 0.43 & 0.02 \\
\end{tabular}
    \caption{Results for evaluation in test bed, across 100 simulated patients.}
    \label{tab:results}
\end{table*}

\bibliography{meier23}

\end{document}